%% file: document.tex
\documentclass[10pt,journal,compsoc]{IEEEtran}
\usepackage[
backend=biber,
style=ieee,
]{biblatex}
\usepackage{tabularx,booktabs,ragged2e,lipsum,microtype,amsmath}
\ifCLASSINFOpdf
   \usepackage[pdftex]{graphicx}
\else
\fi
\usepackage{subfiles}

\begin{document}
%
\title{GEFA: Early Fusion Approach in Drug-Target Affinity Prediction}
%
%
%
%

\author{Tri Minh Nguyen,
        Thin Nguyen,
        Thao Minh Le,
        and Truyen Tran
\IEEEcompsocitemizethanks{\IEEEcompsocthanksitem Tri Minh Nguyen, Thin Nguyen, Thao Minh Le, Truyen Tran are with the Applied Artificial Intelligence Institute, Deakin University, Victoria, Australia.\protect\\
E-mail: (minhtri, thin.nguyen, lethao, truyen.tran)@deakin.edu.au}%
}

%
%

\markboth{}
{Shell \MakeLowercase{\textit{et al.}}: Bare Demo of IEEEtran.cls for Computer Society Journals}
%



\IEEEtitleabstractindextext{%
\begin{abstract}
\subfile{sections/abstract.tex}
\end{abstract}

\begin{IEEEkeywords}
Drug-target binding affinity, Graph neural network, Early fusion, Representation change.
\end{IEEEkeywords}}

\maketitle

\IEEEdisplaynontitleabstractindextext

%
\IEEEpeerreviewmaketitle

\IEEEraisesectionheading{\section{Introduction}\label{sec:introduction}}

%
%
%
%

\subfile{sections/intro.tex}

\section{Related Works}

\subfile{sections/relatedworks.tex}

\section{Proposed Methods}

\subfile{sections/method.tex}

\section{Experiments}

\subfile{sections/exps.tex}

\section{Conclusion}

\subfile{sections/conclusion.tex}


%



\ifCLASSOPTIONcaptionsoff
  \newpage
\fi



%

\printbibliography

%

\begin{IEEEbiography}[{\includegraphics[width=1in,height=1.25in,clip,keepaspectratio]{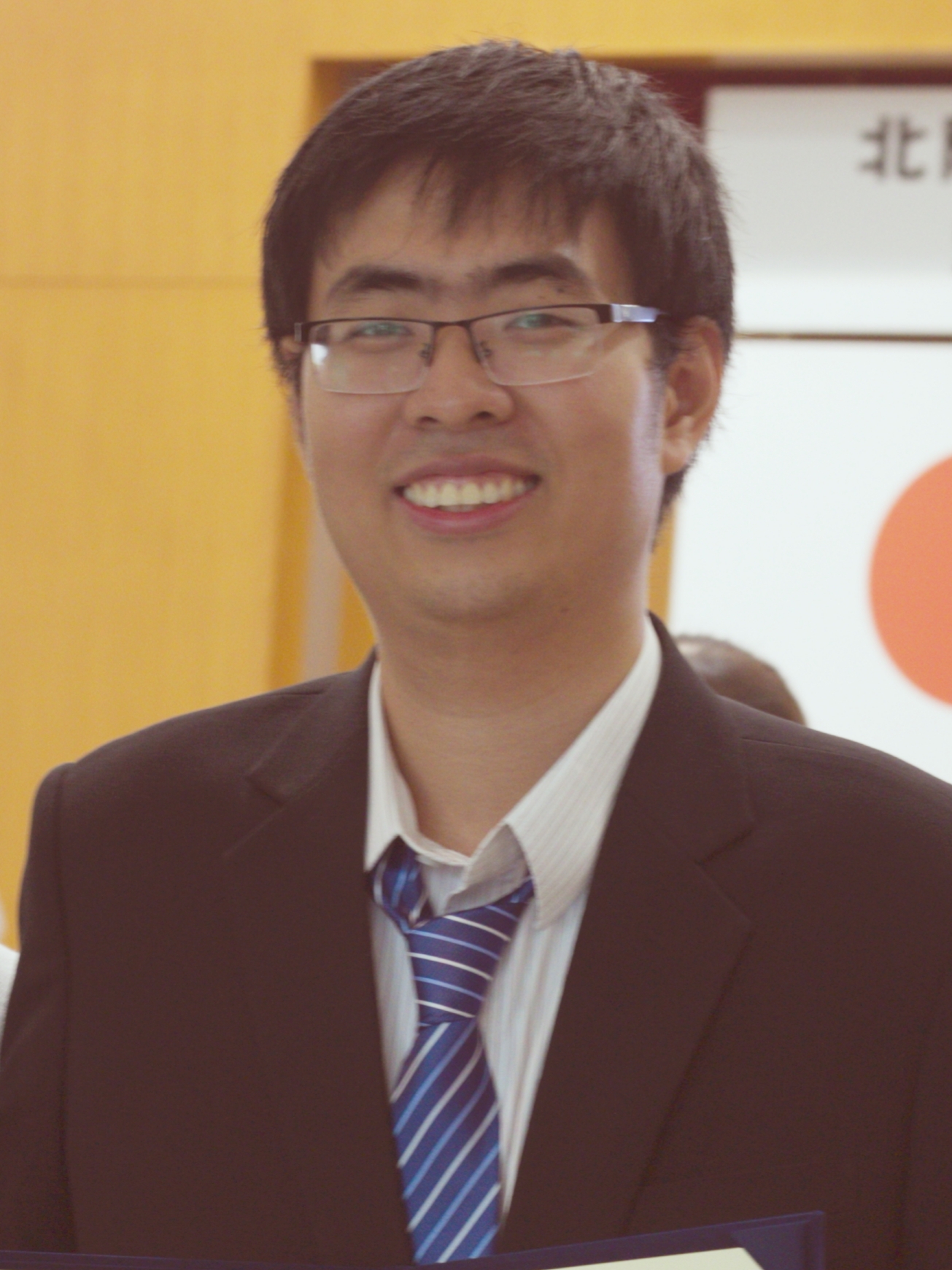}}]
{Tri Minh Nguyen} is a PhD student at Deakin University. His research interest is applying machine learning in studying the protein structure and function. He is exploring the application of energy-based models in protein structure and function.
\end{IEEEbiography}

\begin{IEEEbiography}[{\includegraphics[width=1in,height=1.25in,clip,keepaspectratio]{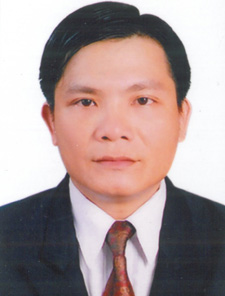}}]
{Thin Nguyen} is a Senior Research Fellow with the Applied Artificial Intelligence Institute (A2I2), Deakin University, Australia. He graduated with a PhD in Computer Science from Curtin University, Australia. His current research topic is inter-disciplinary, bridging large-scale data analytics, pattern recognition, genetics and medicine. His research direction is to develop machine learning methods to discover functional connections between drugs, genes and diseases.
\end{IEEEbiography}

\begin{IEEEbiography}[{\includegraphics[width=1in,height=1.25in,clip,keepaspectratio]{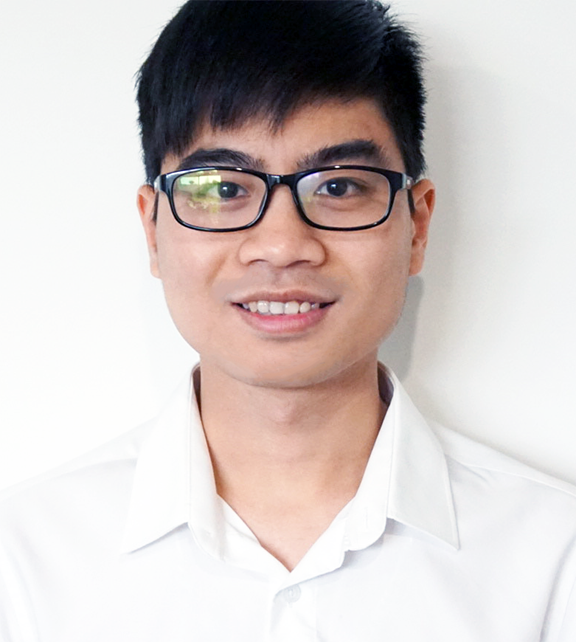}}]
{Thao Minh Le} is a senior PhD student at Applied Artificial Intelligence Institute, Deakin University. His research interests focus on computer vision and its applications, and machine reasoning. He is currently exploring new frontiers in learning and reasoning across different modalities.
\end{IEEEbiography}

\begin{IEEEbiography}[{\includegraphics[width=1in,height=1.25in,clip,keepaspectratio]{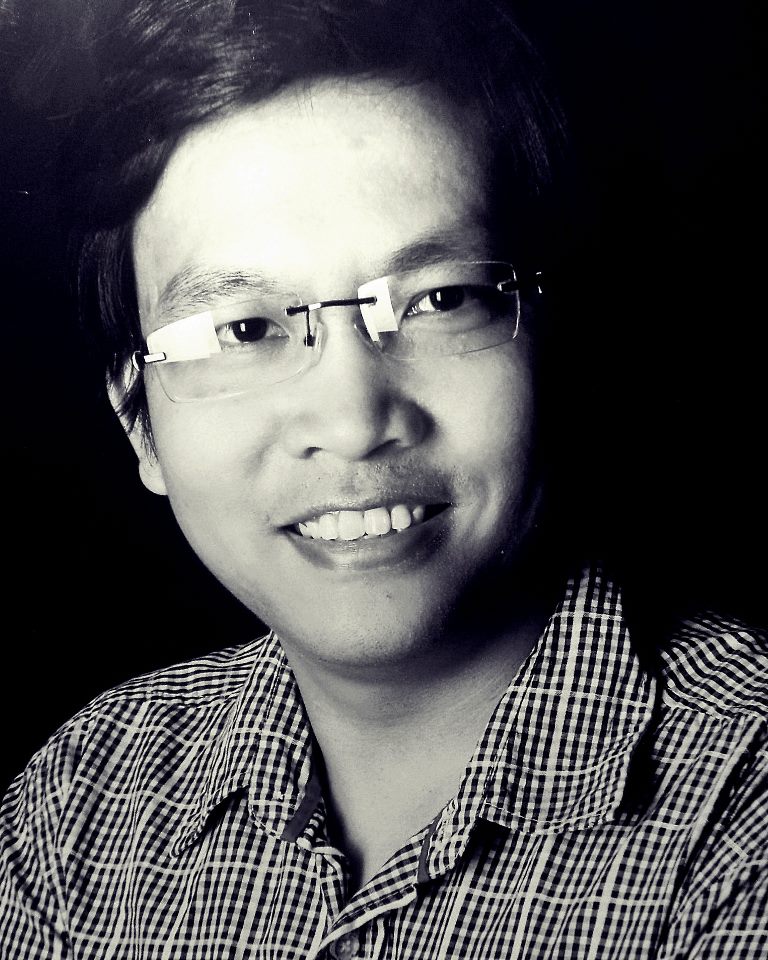}}]
{Truyen Tran} is an Associate Professor at Deakin University. He is member of Applied Artificial Intelligence Institute where he leads the work on deep learning and its application on health, genomics, software and materials science. His other research topics include probabilistic graphical models, recommender systems, learning to rank, anomaly detection, multi-relational databases, model stability, and mixed-type analysis.
\end{IEEEbiography}




\end{document}

%% file: sections/abstract.tex
Predicting the interaction between a compound and a target is crucial
for rapid drug repurposing. Deep learning has been successfully applied
in drug-target affinity (DTA) problem. However, previous deep learning-based
methods ignore modeling the direct interactions between drug and protein
residues. This would lead to inaccurate learning of target representation
which may change due to the drug binding effects. In addition, previous
DTA methods learn protein representation solely based on a small number
of protein sequences in DTA datasets while neglecting the use of proteins
outside of the DTA datasets. We propose GEFA (Graph Early Fusion Affinity),
a novel graph-in-graph neural network with attention mechanism to
address the changes in target representation because of the binding
effects. Specifically, a drug is modeled as a graph of atoms, which
then serves as a node in a larger graph of residues-drug complex.
The resulting model is an expressive deep nested graph neural network.
We also use pre-trained protein representation powered by the recent
effort of learning contextualized protein representation. The experiments
are conducted under different settings to evaluate scenarios such
as novel drugs or targets. The results demonstrate the effectiveness
of the pre-trained protein embedding and the advantages our GEFA in
modeling the nested graph for drug-target interaction.

%% file: sections/intro.tex
\IEEEPARstart{P}{redicting} drug-target binding affinity (DTA prediction) is crucial
in new drug development as well as drug repurposing \cite{thafar2019comparison}.
The gold standard to determine the binding affinity is by experimental
assays but this is prohibitively expensive as a rapid screening tool
as there are over 100 million drug-like compounds \cite{kim2019pubchem}
and over 5000 potential protein targets \cite{gilson2016bindingdb}.
Therefore, it is necessary to have alternative computational methods
using simulation or machine learning to predict the binding affinity
of novel drug-target pairs. Machine learning methods are particularly
attractive because they offer cheap and fast alternatives with reasonable
performance thanks to the large DTA databases \cite{gilson2016bindingdb}
that we can leverage on.

With the advance of machine learning, many computational prediction
methods \cite{cichonska2017computational,cichonska2018learning,ozturk2018deepdta,nguyen2019graphdta}
have been proposed to tackling DTA. In recent works, the protein is
typically represented as a string of amino acids denoted by letters
\cite{ozturk2018deepdta,nguyen2019graphdta,ozturk2019widedta}. The
drawback of using protein sequence is that it can not represent the
3D structure of the protein which is crucial information for determining
the binding affinity between protein and drug in practice\cite{li2016identification}.
However, obtaining the high-resolution 3D structure is a challenging
task. A more practical solution is using the 2D residue contact maps
to represent tertiary protein structure. These maps can now be determined
with reasonable accuracy from deep learning powered algorithms \cite{wang2018analysis,senior2020improved}.

The contact map can be naturally modeled using recent advances in
deep learning known as graph neural networks (GNN). Here each residue
is represented as a node in the graph, and a contact between two residues
as an edge. Each node is first embedded with a feature vector. The
GNN operates by iteratively sending messages between nodes, effectively
refining the node representation by collecting information from its
neighbors and distant nodes. Then the protein representation can be
pooled from its residues.

Previous deep learning-based DTA prediction methods \cite{ozturk2018deepdta,nguyen2019graphdta,ozturk2019widedta,zheng2020predicting}
often use the late fusion approach. The late fusion approach extracts
drug and target representation separately then predicts the binding
affinity from the combined representation at the very end of the process.
However, this practice ignores the fact that the binding occurs at
a pocket rather than the whole protein. The pocket is a small convex
cave in the 3D structure of the protein that encourages stable binding
of a drug to the protein. Once the drug binds, it changes the protein
functions to have pharmaceutical effects, hence it can also change
the protein structure \cite{teague2003implications}, hence its representation.
As the late fusion approach only combines drug and target representation
extracted from the input, the change in protein representation due
to the binding process is not addressed. In addition, the model assumes
non site-specific binding, making it difficult to assign the credit
to the sites that interact. It can also result in slower learning
rate, and less interpretable prediction.

To address target protein representation change, we propose an early-fusion-based
approach. Initially, we extract representation feature for a given
drug molecule from its drug graph structure. Then, the drug representation
is integrated into the protein graph structure before the protein
representation learning phrase. This is basically a graph structure
nested inside another graph structure. This graph-in-graph neural
network design allows the model to learn changes in protein representation
caused by the binding process with the drug molecule.

Previous works \cite{ozturk2019widedta,ozturk2018deepdta,nguyen2019graphdta,jiang2020drug}
normally use one-hot encoding to vectorize residues. This conventional
approach fails to embed the contextual dependencies between residues
as well as not being able to make use of unlabeled protein sequences.
Recent advances in natural language processing allow to learn contextual
embedding from massive unlabeled data which are the current state-of-the-art
on many tasks \cite{devlin2019bert}. Therefore, we take advantage
of the power of the protein embedding features learned by a protein
language modeling on a large collection of protein sequences, including
proteins that are not available in DTA datasets, to represent the
residues in a given target protein. In this work, we refer target
proteins whose the binding affinities available in the DTA datasets
as labeled ones while proteins that do not exist in the DTA datasets
as unlabeled proteins.

In summary, the contribution of our work is two-fold. First, we combine
the protein sequence embedding feature and protein contact map to
build the graph representation of a target protein. Second, in order
to reflect the target representation change during the binding process,
we propose a so-called Graph Early Fusion for binding Affinity prediction
(GEFA) for more accurate biological modeling. We demonstrate the effects
of the GEFA on Davis dataset \cite{davis2011comprehensive} where
it has shown superior performance against previous studies on different
settings. Our Python implementation and data are publicly available
at \url{https://github.com/ngminhtri0394/GEFA}.

%% file: sections/relatedworks.tex
\subsection{Drug Re-purposing as an Alternative Medication for Novel Disease}

Drug re-purposing \cite{langedijk2015drug} is the process of identifying
well-established medications for the novel target disease. The advantages
of this drug re-purposing over developing a completely novel drug
are lower risk and fast-track development \cite{xue2018review}. The
process of drug re-purposing consists of three key steps: identifying
the candidate molecules given the target disease, drug effect assessment
in the preclinical trial, and effectiveness assessment in clinical
trial \cite{pushpakom2019drug}. The first step, hypothesis generation,
is critical as it decides the success of the whole process. Advanced
computational approaches are used for hypothesis generation. Computational
approaches in drug re-purposing can be categorized into six groups
\cite{pushpakom2019drug}: genetic association \cite{grover2015novel,sanseau2012use},
pathway pathing \cite{iorio2010identification,iorio2013network,smith2012identification},
retrospective clinical analysis \cite{hurle2013computational,jensen2012mining,paik2015repurpose},
novel data sources, signature matching \cite{dudley2011exploiting,iorio2013transcriptional,sirota2011discovery},
molecular docking \cite{kitchen2004docking,dakshanamurthy2012predicting,cooke2015structures}. 

\subsection{Drug-Target Binding Affinity Prediction Problem}

Drug-target binding affinity indicates the strength of the binding
force between the target protein and its ligand (drug or inhibitor)
\cite{ma2018overview}. The drug-target binding affinity prediction
problem is a regression task predicting the value of the binding force.
The binding strength is measured by the equilibrium dissociation constant
($K_{D}$). A smaller $K_{D}$ value indicates a stronger binding
affinity between protein and ligand \cite{ma2018overview}. There
are two main approaches: structural approach and non-structural approach
\cite{thafar2019comparison}. Structural methods utilize the 3D structure
of protein and ligands to run the interaction simulation between protein
and ligand. On the other hand, the non-structural approach relies
on ligand and protein features such as sequence, hydrophobic, similarity
or other alternative structural information.

\subsubsection{Structural Approach}

The structure-based approach involves molecular docking, predicting
the three-dimensional structure of the target-ligand complex. In molecular
docking, there are a large number of target-ligand complex conformations.
The conformations are evaluated by the scoring function. Based on
the scoring function types, the structural approach can be categories
into three groups \cite{thafar2019comparison}: classical scoring
function method \cite{meng1992automated,jorgensen1983comparison,pullman2013intermolecular,raha2007role},
machine learning scoring function method \cite{kundu2018machine},
and deep learning scoring function method \cite{stepniewska2018development,gomes2017atomic}.

\subsection{Non-structural Approach}

The non-structural approach solves the binding affinity regression
task without the accurate 3D structure of the target. Instead of using
the 3D coordinate of target and drug atom. The non-structural approach
relies on the drug-drug, target-target similarity, target and drug
atom sequence, and other alternative structural information such as
contact map or secondary structure.

\subsubsection{Representation of Protein in Non-structural Approach}

The simplest way to represent a protein chain is by a string of letters.
Twenty alphabet characters are used to encode twenty types of amino
acids. The torsion angle of each amino acid in the protein sequence
is represented as a pair of $\phi$ and $\psi$ angle. Amino acids
in protein sequence have one of three main types of secondary structure:
$\alpha$ helix, $\beta$ pleated sheet, and coil. Therefore, three
alphabet characters are used to encode the secondary structure. The
interaction between two non-adjacency residues is represented as the
contact map or distance map. Two non-adjacency residues are in contact
if their distance is less than 8 ${\AA}$. The contact
map is the binary map showing whether two residues in the protein
chain are in contact.

\subsubsection{Kernel Based Approach (KronRLS)}

KronRLS \cite{cichonska2017computational,cichonska2018learning} uses
the kernel-based approach to construct the similarity between drugs
or between target proteins. In KronRLS, a kernel is a function measuring
the similarity between two molecules. The regularized least squares
regression (RLS) framework is used to predict the binding affinity
values. 

\subsubsection{Similarity Matching Approach (SimBoost)}

SimBoost \cite{he2017simboost} constructs features for each drug,
target, and drug-target pair from the similarity among drugs and targets.
Then similarity-based features are used as input in a gradient boosting
machine to predict the continuous value of binding affinity of the
drug-target couples. 

\subsubsection{Deep Learning Based Methods}

DeepDTA \cite{ozturk2018deepdta} predicts the binding affinity from
the 1D representation of protein and drug. WideDTA \cite{ozturk2019widedta}
is an extension of DeepDTA. Protein is represented not only in sequence
but also in motif and domain. The drug is represented in SMILES and
Ligand Maximum Common Substructures. The method's drawback is using
sequence to represent drugs and targets. However, the structural information
of the drug and the target plays important roles in the drug-target
interaction. Instead of using 1D representation for drug, GraphDTA
\cite{nguyen2019graphdta} uses graph to express the interaction between
atoms of the molecules. This allows modeling the interaction between
any two atoms within the drug molecules. However, the protein is still
represented as sequence which limits the model performance. PADME
\cite{feng2018padme} uses graph structure to represent the compound
and Protein Sequence Information (PSC descriptors) to represent the
target. The PSC descriptors contains richer information than the sequence.
However, the model performance may be limited without the structural
information. DrugVQA \cite{zheng2020predicting} uses distance map
to represent the protein. Sequential self-attention is used to learn
which parts of the protein interact with the ligand. Multi-head self-attention
is used to learn which atoms in drugs have high contribution to the
drug-target interaction. However, DrugVQA is a supervised learning
method without any pretraining on targets or drugs. Therefore, it
may not cope well with a novel drug or protein. Graph-CNN \cite{torng2019graph}
pretrains the protein pocket graph autoencoder by minimizing representation
difference. The binding interaction model has protein pocket graph
and 2D molecular graph as the inputs. The unsupervised learning helps
the model to overcome the limited pocket graph training data. However,
Graph-CNN requires the protein pockets as target representation which
may limit its general usage. DGraphDTA \cite{jiang2020drug} uses
contact map to build protein graph structure with PSSM, one-hot encoding,
and residue properties as node features. This allows the model to
obtain an accurate protein representation. However, DGraphDTA ignores
the representation change of the target caused by conformation change.
As a result, the target representation learned by the model may be
inaccurate.

%% file: sections/method.tex
The task of drug-target binding affinity (DTA) problem is to predict
the binding affinity A between a target protein P and drug compound
D. Mathematically, the problem is formulated as a regression task:
\begin{equation}
A=\mathcal{F}_{\theta}(P,D),
\end{equation}
where $\theta$ is model parameters of predicting function $\mathcal{F}$.

In this section, we present details of our approach to solve DTA.
In Sec. \ref{graph_prot} explains the feature representation of target
protein P, followed by the feature representation of drug compound
D in Sec. \ref{graph_drug}. The main contribution of ours is presented
in Sec. \ref{gefa_method} where we aim at reflecting the changes
in the target protein representation due to the conformation change.

\subsection{Graph Representation of Protein}

\label{graph_prot} Previous methods applying deep learning to DTA
\cite{ozturk2018deepdta,ozturk2019widedta,nguyen2019graphdta} use
target protein sequence whose residues are embedded into vector space,
often with one-hot encoding. Subsequently, an encoder of multiple
1D convolution layers are used to obtain a final presentation of the
protein sequence. This approach only makes use of the primary structure
information of the target protein sequence. However, the tertiary
information is important for the drug-target interaction \cite{li2016identification}.
Even though the 3D representation can precisely describe the tertiary
structure, obtaining 3D geometrical structure information of the protein
is time-consuming and challenging. This is even impossible for some
protein types using NMR or X-ray crystallography \cite{lacapere2007determining}.
To balance the complexity against efficiency, we utilize the 2D contact
map information as the representation of the tertiary structure to
represent the protein graph structure.

Given that we have the 2D tertiary structure information, each target
protein is considered as a graph structure where nodes are residues
in the protein sequence. Previous studies \cite{nguyen2019graphdta,ozturk2018deepdta}
use one-hot encoding to vectorize each residue in a protein sequence.
While using one-hot encoding is convenient, it does not tell the similarity
between elements but consider them equidistant from each other. In
addition, this choice of representation limits the learning capability
as it fails to leverage the contextual dependencies between residues,
which can be estimated through unsupervised learning. In effect, the
learning signals only come from the limited number of target proteins
in a particular dataset. In fact, there are only 7605 target proteins
\cite{gilson2016bindingdb} being used for the task of drug-target
binding affinity (DTA) prediction while the majority of proteins are
not being used. It has been estimated that there exist over 188 million
sequences of unlabeled target protein sequences\cite{uniprot2019uniprot}. 

Leveraging this rich set of unlabeled proteins, we utilize state of
the art language modeling methods to learn contextual residue embedding
representation. We use embedded representation learnt from a large
collection of unlabeled protein sequences provided by TAPE \cite{rao2019evaluating}
instead of one-hot encoding to represent each node in the protein
graph. To be more specific, it relies on the latest advance of self-supervised
learning which is widely studied in natural language processing. Self-supervised
learning is a subset of unsupervised learning where the supervision
is derived from the data itself \cite{jing2020self}. Generally, a
part of data is withheld and the model is trained to predict it. Language
modeling is a common case of self-supervised learning in natural language
processing in which it learns the representation of sequence using
pretext tasks such as predicting missing token or the next token in
sequence \cite{devlin2019bert}. TAPE is a variant of language model
for protein representation in particular. Subsequently, given a protein
sequence of $L$ residues, the node features of the protein graph
is a set $\mathcal{V}_{p}=\left\{ v_{i}\mid v_{i}\in R{}^{h}\right\} _{i=1}^{L}$,
where $h$ is the length of the embedding vector $v_{i}$ provided
by TAPE. Each $v_{i}$ is contextual, that is residues occur in the
context of surrounding residues. Therefore, the structural information
is implicitly encoded into the embedding.

We use secondary structure as it decides the backbone shape of the
target protein which also contributes to the shape of the binding
site and overall structure. For each residue, the secondary structure
feature is represented as the probability of three secondary structure
type $\alpha$ helix, $\beta$ pleated sheet, and coil.

Solvent accessibility indicates the level of interaction between residues
and drug molecule. Solvent accessibility is divided into three classes:
buried (pACC from 0 to 10), medium (pACC from 11 to 40), and exposed
(pACC from 41 to 100). Residues buried inside the protein core are
less likely to interact with the drug molecule while exposed residues
are more likely to interact with drug molecule. Eventually, the combination
of embedding vector extracted by TAPE, secondary structure feature
vector, and solvent accessibility feature vector are used to represent
node features of residues in a target protein graph.

The contact map information provides the contacts between any two
residue nodes in a protein graph. The sequence information is also
retained in the graph structure in form of edges linking any two nodes
of adjacency residues in the protein sequence. In practice, the contact
map and sequence information are stored as an adjacency matrix $\mathcal{A}_{p}$.
In the rest of this paper, we denote the protein graph as $\mathcal{G}_{p}=(\mathcal{V}_{p},\mathcal{A}_{p})$,
where $\mathcal{V}_{p}$ are residue nodes in the protein chain.

\subsection{Graph Representation of Drugs Compounds}

\label{graph_drug} The input drug compound is in the SMILES format.
In the graph representation of molecule, atoms are nodes while the
bonds between atoms are edges. The node feature consists of five properties:
atom symbol, atom degree which is the total number of bonded atom
neighbors, the number of hydrogens, implicit value of the atom, and
whether if the atom is aromatic. These features are concatenated to
form a multi-dimensional feature. The edges are expressed by an adjacency
list indicating if there are bonds between any two atoms in the compounds.
As the bonds are symmetric, a drug compound graph is a bidirectional
graph. In the later use of graph representation of a drug compound,
we refer it as $\mathcal{G}_{d}=(\mathcal{V}_{d},\mathcal{A}_{d})$,
where $\mathcal{V}_{d}$ is atom features and $\mathcal{A}_{d}$ is
bonds between atoms.

\subsection{Graph Early Fusion for binding Affinity prediction (GEFA)}

\label{gefa_method} 
\begin{figure*}
\centering{}\includegraphics[width=0.9\linewidth]{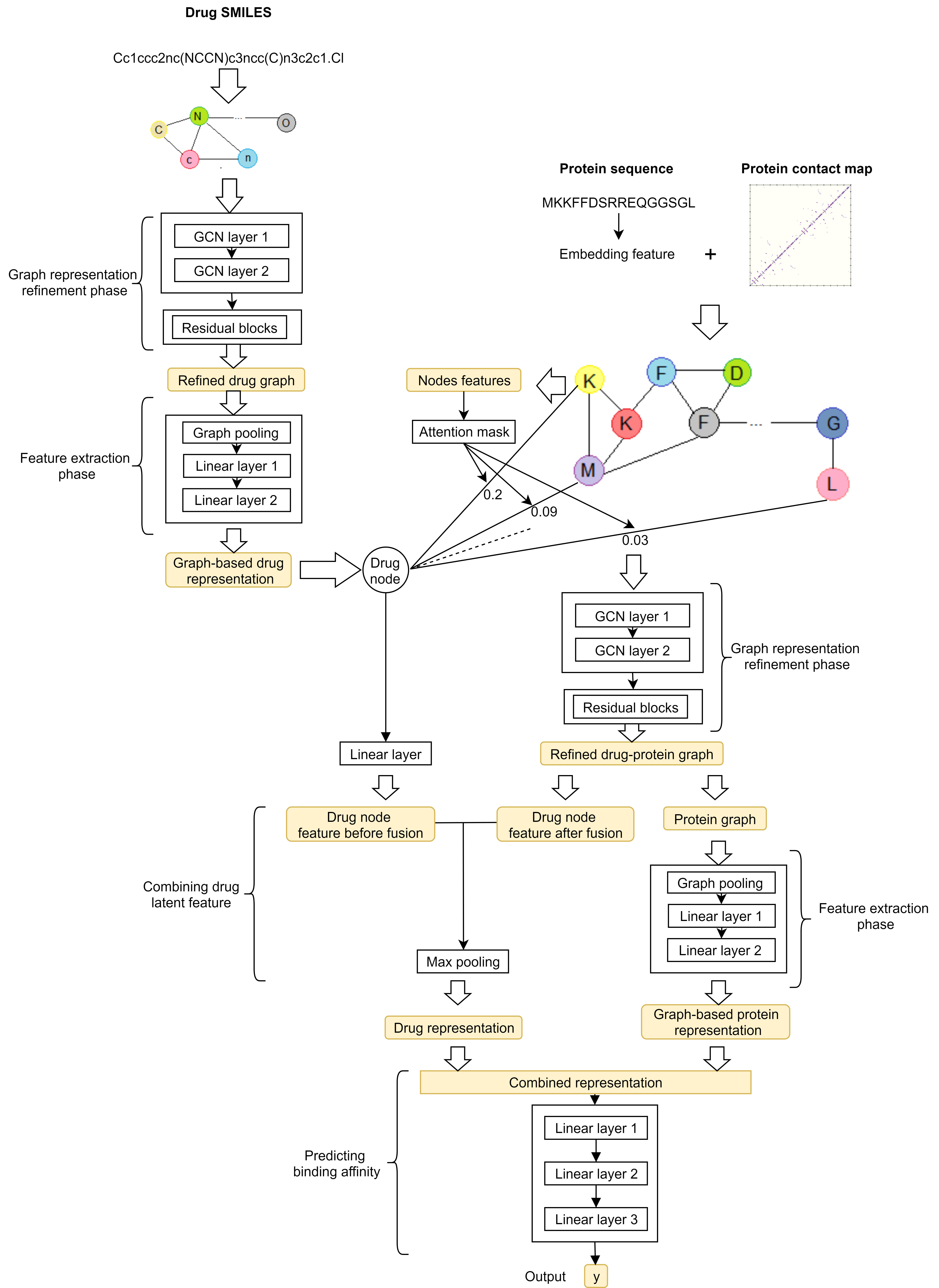}\caption{Illustration of Graph Early Fusion for binding Affinity prediction
(GEFA). The GEFA takes as input the graph representations of a drug
molecule and a protein target. We first use graph convolution network
(GCN) for the drug graph feature refinement before applying a max
pooling operator to obtain the estimated representation of the entire
drug graph. The drug estimated vector is then used as an additional
node to the protein graph, establishing graph-in-graph representation
across domains. Information retrieved from the drug-protein graph
along with the drug representation are finally used as the input for
predicting the binding affinities between the given drug molecule
and the protein target.}
\label{fig:early_fusion} 
\end{figure*}

The overall architecture of our proposed method
is presented in Fig.\ref{fig:early_fusion}. Our GEFA takes as input
the graph structure of drug $\mathcal{G}_{d}$ and the graph structure
of target $\mathcal{G}_{p}$ and outputs the prediction of binding
affinity. We use Graph Convolutional Network (GCN) \cite{kipf2016semi}
for graph representation. In addition, we also make use of the well-known
residual skip connection trick to make use of very deep GCNs.

\subsubsection{Graph Convolutional Network}

GCN is a convolutional network designed specifically for graph-structured
signals. The goal is learn the node-level representation from a given
input graph $\mathcal{G}=(\mathcal{X},\mathcal{A})$ where $\mathcal{X}$
is node feature matrix of $N$ nodes and $\mathcal{A}\in R^{N\times N}$
is the adjacency matrix that describes the graph structure. Let $W^{l}$
be the weight matrix at $l$-th layer, the graph convolution operation
is defined by:

\begin{align}
H^{1} & =\mathcal{X},\\
H^{l} & =\sigma\left(\tilde{D}^{-\frac{1}{2}}\tilde{\mathcal{A}}\tilde{D}^{-\frac{1}{2}}H^{l-1}W^{l-1}\right),\label{eq:gcn_layer}
\end{align}
where $\mathcal{\tilde{A}=\mathcal{A}+I}$ is the adjacency matrix
with self-loop in each node. $\mathcal{I}$ is the identity matrix
and $\tilde{D}=\sum_{j}\mathcal{\tilde{A}}_{ij}$ and $\sigma$ is
a non-linear function which is a ReLU \cite{agarap2018deep} in our
later experiments.

\subsubsection{Deeper GCN with Residual Blocks}

In general, a deeper model can generalize better and more compact
than shallow networks\cite{Goodfellow-et-al-2016}. However, stacking
the vanilla GCN often suffers from the problem of gradient vanishing
and numerical instability as a consequence of matrix multiplication
in Eq. \ref{eq:gcn_layer}. To mitigate this problem, we use the GCN
with residual skip-connection proposed in \cite{le2020dynamic}. Similar
to the effect of the residual block in the well-known CNN \cite{he2016deep},
skip connection in GCN helps to create more direct gradient flow,
hence, allows to go deeper with more convolution layers. Mathematically,
the graph convolution operation is given by:
\begin{align}
H^{1}= & \mathcal{X},\label{eq:gcn_res_1}\\
F^{l}(H^{l-1})= & W_{2}^{l-1}\sigma(\mathcal{A}H^{l-1}W_{1}^{l-1}+b^{l-1}),\label{eq:gcn_res_2}\\
H^{l}= & \sigma(H^{l-1}+F^{l}(H^{l-1})),\label{eq:gcn_res_3}
\end{align}
where $W_{1},W_{2}$ are learnable weight matrices, $l$ is layer-wise
index and $\sigma$ is a non-linear function which is a ReLU activation
function.

\subsubsection{Graph-Graph Integration with Early Fusion}

To reflect the changes in representation of a target protein due to
the interaction between drug molecule and protein, we propose Graph
Early Fusion for binding Affinity (GEFA), a method for migrating the
drug molecule graph $\mathcal{G}_{d}=(\mathcal{V}_{d},\mathcal{A}_{d})$
into the protein graph $\mathcal{G}_{p}=(\mathcal{V}_{p},\mathcal{A}_{p})$
via a self-attention mechanism.

We first refine node representations in the drug graph with a two-layers
GCN as in Eq. \ref{eq:gcn_layer} and residual blocks as in Eq. \ref{eq:gcn_res_1},
\ref{eq:gcn_res_2}, \ref{eq:gcn_res_3}. Let $\mathcal{V}_{d}^{\prime}=\{v_{i}^{\prime}\mid v_{i}^{\prime}\in R^{h_{1}}\}_{i=1}^{S}$
as node features of the drug graph after GCN, where $S$ is number
of nodes in the drug graph. Note that $v_{i}^{\prime}$ contains aggregating
information from its neighbors so we simply use the largest estimated
representation of the refined drug graph as the representation of
the entire graph. This is easily obtained by a max pooling operation
followed by two linear layers for feature projection: 
\begin{equation}
v_{\textrm{max}}^{\prime}=\textrm{MaxPool}(\mathcal{V}_{d}^{\prime}),
\end{equation}
\begin{equation}
x_{d}=(W_{0}v_{\textrm{max}}^{\prime}+b_{0})W_{1}+b_{1}.
\end{equation}
We call the resulted vector $x_{d}\in R^{h_{1}}$ as the drug molecules
node, where $h_{1}$ is dimension of $x_{d}$.

We now explain how we integrate the drug molecules node $x_{d}$ into
the protein graph $\mathcal{G}_{p}=(\mathcal{V}_{p},\mathcal{A}_{p})$
which is the main contribution of our work. The key idea is to use
the drug node $x_{d}$ as an additional node that binds to the target
graph $\mathcal{G}_{p}$. The edges connecting the drug node and residue
nodes in the protein graph indicate the interaction between residues
and drug molecule as well as the binding site. Since not every residue
contributes equally to the binding affinity, the edge weights indicate
the level of interactions of each residue with the drug molecule.
To learn the level of contribution, we utilize a self-attention mechanism
driven by the residue features $\mathcal{V}_{p}=\{v_{i}\mid v_{i}\in R^{h_{2}}\}_{i=1}^{L}$,
recalling that $L$ is the length of the protein sequence. The self-attention
mechanism is motivated from the fact that the binding site of the
protein depends on the protein structure. In the other word, the attention
weights tell which residues are more likely to participate in the
binding process. Mathematically, the attention weights are given by:
\begin{equation}
\alpha_{i}=\textrm{sotfmax}(W_{2}\textrm{tanh}(W_{1}v_{i})),
\end{equation}
where $\sum_{i=1}^{L}\alpha_{i}=1,v_{i}$ is the $i$-th residue feature,
and $W_{1}$ and $W_{2}$ are the learnable parameters.

Given the drug node $x_{d}$ and its connections to residues in the
target protein graph $\mathcal{G}_{p}$ denoted by $\{\alpha_{i}\}_{i=1}^{L}$,
we now construct a cross-domain graph $\mathcal{G}_{pd}=\{\mathcal{V}_{pd},\mathcal{A}_{pd}\}$
where $\mathcal{V}_{pd}=\{\mathcal{V}_{p},x_{d}\}$ and $\mathcal{A}_{pd}=\{\mathcal{A}_{p},\{\alpha_{i}\}_{i=1}^{L}\}$.
Similar to what we have done with the drug graph earlier, we employ
a two-layers GCN followed by residual blocks to refine the node representations
of the drug-protein graph $\mathcal{G}_{pd}$.

Before performing the graph feature extraction, the drug node after
fusion $v_{dp}^{\prime}$ in the refined nodes $\mathcal{V}_{pd}^{\prime}$
by GCNs is taken out from the protein graph to ensure that the graph
feature only contains residues nodes. Eventually, we extract the latent
representation of the protein graph with a global max pooling operator
followed by a two-layer linear network: 
\begin{equation}
v_{\textrm{max}}^{\prime}=\textrm{MaxPool}(\mathcal{V}_{p}^{\prime}),
\end{equation}
\begin{equation}
v_{pd}=(W_{0}v_{\textrm{max}}^{\prime}+b_{0})W_{1}+b_{1},
\end{equation}
where $\mathcal{V}_{p}^{\prime}=\{\mathcal{V}_{pd}^{\prime}\setminus v_{dp}^{\prime}\}$
is the node representations of the protein graph after removing the
drug node.

At the same time, the drug latent vector $x_{d}$ is transformed into
the same dimensional space with $v_{dp}^{\prime}$ via a linear transformation.
We further obtain the final representation of drug by combining these
two features with a simple concatenation operator. 
\begin{equation}
v_{db}=[x_{d};v_{dp}^{\prime}],
\end{equation}
where $[\ ;]$ denotes the concatenation operation of two vectors.
A max-pooling operation is then performed along the channel dimension
to obtain the combined drug representation. 
\begin{equation}
v_{dc}=\textrm{MaxPool}(v_{db}).
\end{equation}
Consequently, the drug vector $v_{dc}$ and the protein latent vector
$v_{pd}$ are concatenated and finally fed into a predictor of three
fully connected layers to predict the binding affinities.

We wish to hypothesize that our early fusion approach with self-attention
has two benefits. First, the early fusion approach explicitly models
interactions between the drug graph and the target protein graph.
Second, the self-attention allows the learning model to be more interpretable
by showing which residues interact with drug molecules and how much
they contribute to the binding process. We will back these in our
later experiments.

%% file: sections/exps.tex
We evaluate our proposed model GEFA on Davis dataset\cite{davis2011comprehensive}
and compare against a late fusion baseline as well as state-of-the-art
methods including GCNConvNet \cite{nguyen2019graphdta}, GINConvNet
\cite{nguyen2019graphdta}, DGraphDTA \cite{jiang2020drug}. Among
those methods, GCNConvNet and GINConvNet use protein sequence and
drug molecule graph as the input while DGraphDTA uses a protein graph
built from contact and drug molecule as the input. We present the
qualitative results in Sec. \ref{subsec:Quantitative-Experiments}
and further provide analysis of our proposed model via extensive ablation
studies in Sec.\ref{ablation}.

\subsection{Quantitative Experiments\label{subsec:Quantitative-Experiments}}

\subsubsection{Dataset}

Davis dataset consists of binding affinity information between 72
drugs and 442 targets. The binding affinity between drug and target
is measured by $K_{D}$ (kinase dissociation constant) value \cite{davis2011comprehensive}.
For the Davis dataset, the drug SMILES sequence of 68 drugs and the
target protein sequence of 442 targets from DeepDTA\cite{ozturk2018deepdta}
training/test set are used in our experiments.

There are four experiments settings for four scenarios. The first
experiment setting is the warm setting where both protein and drug
are known to the model. In this case, every protein and drug appear
in training, validation, and test set.

The second experiment is cold-target where proteins are unknown to
the model and drugs are known to the model. This setting replicates
the scenarios of drug repurposing for a novel target. In this case,
each unique protein sequence only appears in training, validation,
or test set. As targets in the cold-target setting are required to
be unique in both train, validation, and test sets, targets having
the same sequence are filtered out which results in 361 targets. The
361 targets are split at 0.8/0.2 ratio for training-validation/testing.
Then the training set is split at 0.8/0.2 ratio for training/validation.

The third experiment setting is cold-drug where proteins are known
to the model and drugs are unknown to the model. In this case, a unique
drug only appears in training, validation, or test set. We conduct
the same splitting procedure in cold-target but applying for drugs.

Finally, the last experiment setting is cold-drug-target where both
drugs and proteins are unknown to the model. This scenario is the
case of novel drugs for the novel target. In this setting, we conduct
the splitting procedure for both drug and target to ensure training,
validation, and testing set do not share any common drug or target. 

\subsubsection{Implementation Details}

Our methods are implemented using Pytorch. The protein sequence embedding
features are extracted using TAPE-Protein \cite{tape2019}. The contact
map is predicted by RaptorX \cite{wang2017accurate}. The TAPE-Protein
uses BERT language modelling \cite{devlin2019bert}. The output of
TAPE-Protein embedding features extraction is a embedding vector size
768. The graph convolution network uses the Pytorch geometric library
\cite{Fey/Lenssen/2019}. The model are trained on $128$ mini-batch.
The learning rate is $0.0005$ in warm setting. In cold-target, cold-drug,
and cold-drug-target setting, the learning rate is $0.001$ as higher
learning rate helps model to have better generalization and less likely
to overfit. The learning rate decay is used. The learning rate is
reduced by $20\%$ every $40$ epochs without improvement in MSE metric
in the validation set. Adam optimizer is used. The model is trained
in $1000$ epochs.

\subsubsection{Evaluation Metrics}

The models' performances are evaluated using Concordance Index (CI)\cite{gonen2005concordance},
Mean Squared Error (MSE), Root Mean Squared Error (RMSE), Pearson\cite{benesty2009noise},
and Spearman\cite{zwillinger1999crc}. CI measures the quality of
the ranking. CI formula is given as: 
\begin{equation}
CI=\frac{1}{Z}\sum_{d_{i}>d_{j}}h(b_{i}-b_{j}),
\end{equation}
\begin{equation}
h(x)=\begin{cases}
1, & \text{if}\ x>0\\
0.5, & \text{if}\ x=0\\
0, & \text{if}\ x<0
\end{cases},
\end{equation}
where $b_{i}$ is the predicted value for the larger affinity value
$d_{i}$, $b_{j}$ is the predicted value for the smaller affinity
value $d_{j}$, Z is the normalization constant, and $h(x)$ is the
step function.

MSE measures the mean squared error of the predicted values. MSE formula
is given as: 
\begin{equation}
MSE=\frac{1}{N}\sum_{i=1}^{N}(p_{i}-y_{i})^{2},
\end{equation}
where $p_{i}$ is the predicted value and $y_{i}$ is the ground truth.
RMSE is the square root of MSE.

Pearson measures the linear correlation between the predicted value
$p$ and the ground truth $y$: 
\begin{equation}
Pearson=\frac{\phi(p,y)}{\phi(p)\phi(y)},
\end{equation}
where $\phi(p,y)$ is the covariance between the predicted value and
the ground truth.

Spearman measures the rank correlation: 
\begin{equation}
Spearman=1-\frac{6\sum_{i=1}^{n}d_{i}^{2}}{n(n^{2}-1)},
\end{equation}
where $d_{i}$ is the difference between two ranks in the predicted
values and ground truths. 

\subsubsection{Results}

To compare our early fusion approach with the conventional late fusion
approach, we provide a late fusion baseline model. Our late fusion
baseline model (GLFA - Graph Late Fusion for binding Affinity) follows
the convention model in which drug and protein representation are
learned separately. GLFA has graph structures of drug and target as
the input. Both graph structures are processed using the two-layers
GCN and residual blocks in parallel to learn the hidden features.
Then, the latent features are obtained by global max pooling followed
by two linear layers. The latent features from both protein and drug
are concatenated before under-going three fully connected layers.
The output of the final fully connected layers is the binding affinity
value of the input drug and target protein. The late fusion model
is a special case of early fusion where all drug-residues edge weights
are set to 0. The late fusion approach is used as baseline to compare
with our proposed early fusion approach. 
\begin{table}[h]
\caption{The result of quantitative experiments\label{Tab:davis}}
\resizebox{\columnwidth}{!}{{%
\begin{tabular}{llllll}
\toprule 
Architecture & RMSE$\downarrow$ & MSE$\downarrow$ & Pearson$\uparrow$ & Spearman$\uparrow$ & CI $\uparrow$\tabularnewline
\midrule 
\textbf{Warm start setting} &  &  &  &  & \tabularnewline
\hspace{2mm}GCNConvNet \cite{nguyen2019graphdta} & 0.5331 & 0.2842 & 0.8043 & 0.6609 & 0.8649\tabularnewline
\hspace{2mm}GINConvNet \cite{nguyen2019graphdta} & 0.50723 & 0.2573 & 0.8245 & 0.6818 & 0.8785\tabularnewline
\hspace{2mm}DGraphDTA \cite{jiang2020drug} & 0.4917 & 0.2417 & 0.8378 & 0.7001 & 0.8869\tabularnewline
\hspace{2mm}GLFA & 0.4850 & 0.2353 & 0.8412 & \textbf{0.7073} & \textbf{0.8950}\tabularnewline
\hspace{2mm}GEFA & \textbf{0.4775} & \textbf{0.2280} & \textbf{0.8467} & 0.7023 & 0.8927\tabularnewline
\midrule 
\textbf{Cold-target setting} &  &  &  &  & \tabularnewline
\hspace{2mm}GCNConvNet \cite{nguyen2019graphdta} & 0.7071 & 0.5000 & 0.5145 & 0.4316 & 0.7293\tabularnewline
\hspace{2mm}GINConvNet \cite{nguyen2019graphdta} & 0.7144 & 0.5104 & 0.5166 & 0.3904 & 0.7065\tabularnewline
\hspace{2mm}DGraphDTA \cite{jiang2020drug} & 0.6855 & 0.4700 & 0.5597 & 0.4941 & 0.7656\tabularnewline
\hspace{2mm}GLFA & 0.6732 & 0.4531 & 0.5828 & 0.5228 & 0.7802\tabularnewline
\hspace{2mm}GEFA & \textbf{0.6584} & \textbf{0.4335} & \textbf{0.6030} & \textbf{0.5506} & \textbf{0.7951}\tabularnewline
\midrule 
\textbf{Cold-drug setting} &  &  &  &  & \tabularnewline
\hspace{2mm}GCNConvNet \cite{nguyen2019graphdta} & 0.9723 & 0.9454 & 0.3385 & 0.3764 & 0.6784\tabularnewline
\hspace{2mm}GINConvNet \cite{nguyen2019graphdta} & 0.9592 & 0.9200 & 0.3779 & 0.3693 & 0.6758\tabularnewline
\hspace{2mm}DGraphDTA \cite{jiang2020drug} & 0.9583 & 0.9184 & 0.3610 & 0.3150 & 0.5337\tabularnewline
\hspace{2mm}GLFA & 0.9280 & 0.8612 & 0.4023 & 0.3549 & 0.6703\tabularnewline
\hspace{2mm}GEFA & \textbf{0.9202} & \textbf{0.8467} & \textbf{0.4515} & \textbf{0.4320} & \textbf{0.7091}\tabularnewline
\midrule 
\textbf{Cold-drug-target setting} &  &  &  &  & \tabularnewline
\hspace{2mm}GCNConvNet \cite{nguyen2019graphdta} & 1.0632 & 1.1304 & 0.1904 & 0.1698 & 0.5782\tabularnewline
\hspace{2mm}GINConvNet \cite{nguyen2019graphdta} & 1.0651 & 1.1345 & 0.1974 & 0.2763 & 0.6275\tabularnewline
\hspace{2mm}DGraphDTA \cite{jiang2020drug} & 1.0749 & 1.1554 & 0.0228 & 0.1795 & 0.6081\tabularnewline
\hspace{2mm}GLFA & 1.0698 & 1.1444 & \textbf{0.3473} & 0.2901 & 0.6362\tabularnewline
\hspace{2mm}GEFA & \textbf{0.9949} & \textbf{0.9899} & 0.3148 & \textbf{0.2932} & \textbf{0.6390}\tabularnewline
\bottomrule
\end{tabular}}{ } }
\end{table}

We report our late fusion approach, GLFA, and early fusion approach,
GEFA, with previous works in Davis benchmark on four settings in Table
\ref{Tab:davis}. Our proposed method GEFA consistently outperforms
previous works in four settings. Our proposed methods achieve state-of-the-art
performance across all four settings. Between two late fusion based
methods DGraphDTA \cite{jiang2020drug} and GLFA, our proposed GLFA
method also outperforms DGraphDTA. This follows our expectations as
the embedding feature contains richer information than one-hot encoding
and PSSM. This also demonstrates the advantage of using the residual
block.

DgraphDTA \cite{jiang2020drug}, GLFA, and GEFA outperform GINConvNet
in all four settings. GINConvNet and GCNConvNet \cite{nguyen2019graphdta}
only use sequence and CNN to learn the target representation. On the
other hand, DgraphDTA \cite{jiang2020drug}, GLFA, and GEFA use the
graph built from the protein contact map and learn the target representation
using GCN. This demonstrates the advantage of using the graph representation
of the contact map.

Between our two proposed methods GEFA and GLFA, the early fusion method
GEFA shows advantages over late fusion method GLFA. This follows our
expectation as the early fusion allows interactions between drug and
protein graph during the graph representation learning phase for more
accurate latent representation.

We can observe a common trend that all models performances in the
cold-drug setting are lower than performance in the cold-target setting.
The reason is that the number of unique target sequences in Davis
dataset is higher than the number of unique SMILES sequence. Therefore,
models have more target protein samples to learn the feature space
of the protein.

\subsection{Ablation Studies}

\label{ablation} To understand the contribution of each component
to the overall performance in the early fusion GEFA model, we remove
each component from the GEFA model. We conduct the ablation experiment
using the Davis dataset benchmark in the warm setting.

First, we evaluate the usage of the embedding feature by comparing
it with the one-hot encoding.

Second, we evaluate the usage of attention mask as the graph edge.
Instead of using attention as drug-residue edge weight, drug-residue
edges are weighted the same as the residue-residue edges in the target
graph.

Third, we evaluate the usage of the 2-layer GCN and the usage of residual
blocks to refine graph structure. As all residual blocks have shared
weight, this reduces the number of parameters, which may help in the
case of over-fitting.

Fourth, we evaluate the residual blocks. We test three cases: without
residual blocks in both protein and drug graph, without protein graph
residual blocks, and without drug graph residual blocks.

Finally, we compare the drug representation extracted from the drug-protein
fusion graph and drug representation extracted from the drug graph.
Instead of fusing two types of drug features followed by pooling,
we only use one type of drug feature to combine with graph-based protein
representation.

\begin{table}[h]
\caption{The result of experiment warm setting with different component in
Davis dataset in warm setting. The first row shows proposed GEFA with
all components.)\label{Tab:abla}}
\resizebox{\columnwidth}{!}{{%
\begin{tabular}{p{3.5cm}lllll}
\toprule 
Architecture & RMSE$\downarrow$ & MSE$\downarrow$ & Pearson$\uparrow$ & Spearman$\uparrow$ & CI $\uparrow$\tabularnewline
\midrule 
GEFA & \textbf{0.4775} & \textbf{0.2280} & \textbf{0.8467} & 0.7023 & 0.8927\tabularnewline
\midrule 
One-hot encoding & 0.5050 & 0.2551 & 0.8274 & 0.6919 & 0.8837\tabularnewline
\midrule 
W/o attention & 0.4887 & 0.2388 & 0.8392 & 0.7014 & 0.8909\tabularnewline
\midrule 
W/o 2-layer GCN & 0.4844 & 0.2346 & 0.8425 & 0.6933 & 0.887\tabularnewline
\midrule 
\textbf{Residual blocks usage} &  &  &  &  & \tabularnewline
\hspace{2mm} W/o residual blocks & 0.4933 & 0.2434 & 0.8351 & 0.686 & 0.8819\tabularnewline
\hspace{2mm} Drug graph res. blocks & 0.4944 & 0.2444 & 0.8351 & 0.6874 & 0.8828\tabularnewline
\hspace{2mm} Protein graph res. blocks & 0.4873 & 0.2375 & 0.8407 & 0.7045 & 0.8941\tabularnewline
\midrule 
\textbf{Drug representation choice} &  &  &  &  & \tabularnewline
\hspace{2mm} Before fusion rep. & 0.4806 & 0.231 & 0.8448 & \textbf{0.7058} & \textbf{0.8954}\tabularnewline
\hspace{2mm} After fusion rep. & 0.5171 & 0.2673 & 0.8174 & 0.6437 & 0.8558\tabularnewline
\bottomrule
\end{tabular}}{ } }
\end{table}

As shown in Table. \ref{Tab:abla}, model using the protein embedding
feature has an improvement of $11.85\%$ in MSE and $1.00\%$ in CI.
This emphasizes the advantage of using the protein embedding feature
as the graph node feature.

Using attention mask as the drug-residue edge in protein graph gains
improvement of $4.74\%$ in MSE and $0.20\%$ in CI. Without the attention
mask, we assume that the drug molecule interacts with all residues
in target protein equally. However, each residue contributes differently
to the binding process. Therefore, it is reasonable to use self-attention
to learn each residue's contribution level which is used as edge weight
between drug node and residue node.

Using the 2-layers GCN (GEFA in Table. \ref{Tab:abla}) shows improvement
compared to without using 2-layer GCN (w/o using the 2-layers GCN
in Table. \ref{Tab:abla}). Compared to the model using solely residual
blocks as graph refinement (w/o 2-layer GCN in Table. \ref{Tab:abla}),
the model with residual block shows an advantage over the model with
only the 2-layers GCN (w/o graph residual blocks in Table. \ref{Tab:abla}).
Therefore, it can be suggested that residual blocks have the same
or even better learning ability than the 2-layer GCN. Interestingly,
combining both the 2-layers GCN and residual blocks brings the best
result as shown in the full component model GEFA.

The model using residual blocks in both drug and protein graph shows
an advantage over the model without any residual blocks. Model having
residual blocks for both drug graph and protein graph gains $6.73\%$
improvement in MSE over the model without residual blocks. It is interesting
that applying residual blocks only for drug graph slightly decrease
model performance (0.2\% decrease in MSE and 0.1\% decrease in CI).
Adding back residual blocks for protein graph helps the model to gain
$2.43\%$ improvement in MSE. Stacking residual blocks in protein
graph is more crucial in the early fusion approach as it affects not
only protein graph representation but also the drug node representation.

Finally, we compare the drug representation before and after the drug-target
graph fusion. Model using only drug representation before fusion shows
comparable performance while the model using drug representation after
fusion suffers $17.24\%$ performance loss in MSE. This indicates
that drug representation before fusion is more useful than after fusion.
The reason is likely due to message passing in graph neural network.
The drug node info is updated from its neighbor residue nodes. Therefore,
this suggests that the binding process does not bring any signification
change to the ligand latent representation.

\subsection{Error Analysis}

\begin{figure}[h]
\centering{}\includegraphics[width=0.8\columnwidth]{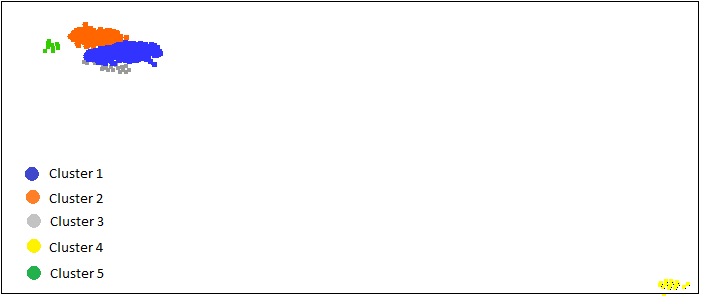}\caption{The distribution of five major target sequence clusters.}
\label{fig:cluster} 
\end{figure}

\begin{figure}[h]
\centering{}\includegraphics[width=0.8\columnwidth]{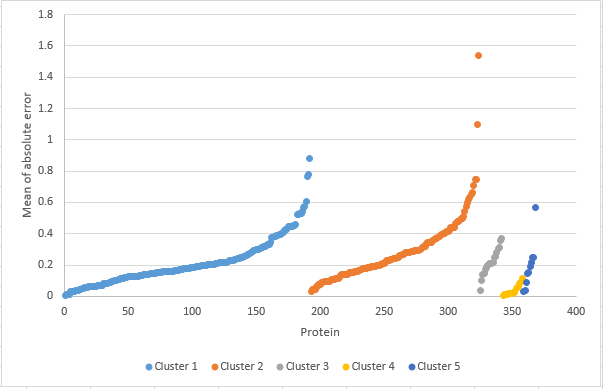}\caption{The average absolute error of each target by five major clusters.}
\label{fig:clustererror} 
\end{figure}

We analyze how protein sequences are clustered and their effect on
model performance. First, we cluster target sequences based on BLAST+
similarities using CLANS \cite{frickey2004clans} convex clustering
on a 2D plane. Five major clusters having more than 10 targets are
chosen to analyze the model performance. The distribution of five
clusters in 2D is shown in Fig. \ref{fig:cluster}. The average absolute
errors of five major clusters are displayed in Fig. \ref{fig:clustererror}
and Table. \ref{Tab:infocluster}. As shown in Fig. \ref{fig:cluster},
cluster 1, 2, 3, and 5 are close to each other in the 2D similarity
distribution space. The average error of cluster 1, 2, 3, and 5 are
roughly the same at $0.2\%$. On the other hand, cluster 4 has a high
distance to other clusters. There is a significant difference in model
performance in cluster 4 compared to other clusters. There are two
possible explanations. First, good performance is from the over-representation
of specific targets in the training set. However, this is not the
case as in all five clusters the training sample/testing sample ratio
is nearly the same at $0.8/0.2$. Second, the difference in performance
is likely due to some specific characteristics of the cluster such
as sequence length, motif, etc.

\begin{table}[h]
\caption{The number of training and testing samples, the number of protein,
and average absolute error of five major sequence clusters.)\label{Tab:infocluster}}
\resizebox{\columnwidth}{!}{{%
\begin{tabular}{p{1cm}p{1cm}p{1cm}p{1cm}p{1cm}p{3cm}}
\toprule 
Cluster & N\textsuperscript{\underline{o}} pairs in testing & N\textsuperscript{\underline{o}} pairs in training & Train/test ratio & N\textsuperscript{\underline{o}} proteins & Average absolute error\tabularnewline
\midrule 
1 & 2159 & 10952 & 0.1971 & 192 & 0.2137 $(\pm0.1492)$\tabularnewline
2 & 1519 & 7455 & 0.2037 & 132 & 0.2882 $(\pm0.2059)$\tabularnewline
3 & 202 & 1022 & 0.1977 & 18 & 0.2189 $(\pm0.0849)$\tabularnewline
4 & 188 & 900 & 0.2089 & 16 & 0.0386 $(\pm0.1166)$\tabularnewline
5 & 117 & 563 & 0.2078 & 10 & 0.1923 $(\pm0.1449)$\tabularnewline
\bottomrule
\end{tabular}}{ } }
\end{table}

We analyze the binding site predicted by the GEFA model in the successful
case and the failed case. We choose two drug-target binding pairs
with 3D structure available having the lowest and highest absolute
error. For the successful case, we analyze the binding between target
MST1 and compound Bosutinib (PubChem CID 5328940) predicted by our
proposed model and compare it with blind docking simulation \cite{biohpc2017achilles}.
The Bosutinib-MST1 pair has predicted $K_{D}$ value at 6.72132 with
the absolute error at 0.0000738.

\begin{figure}[h]
\centering{}\includegraphics[width=0.8\columnwidth]{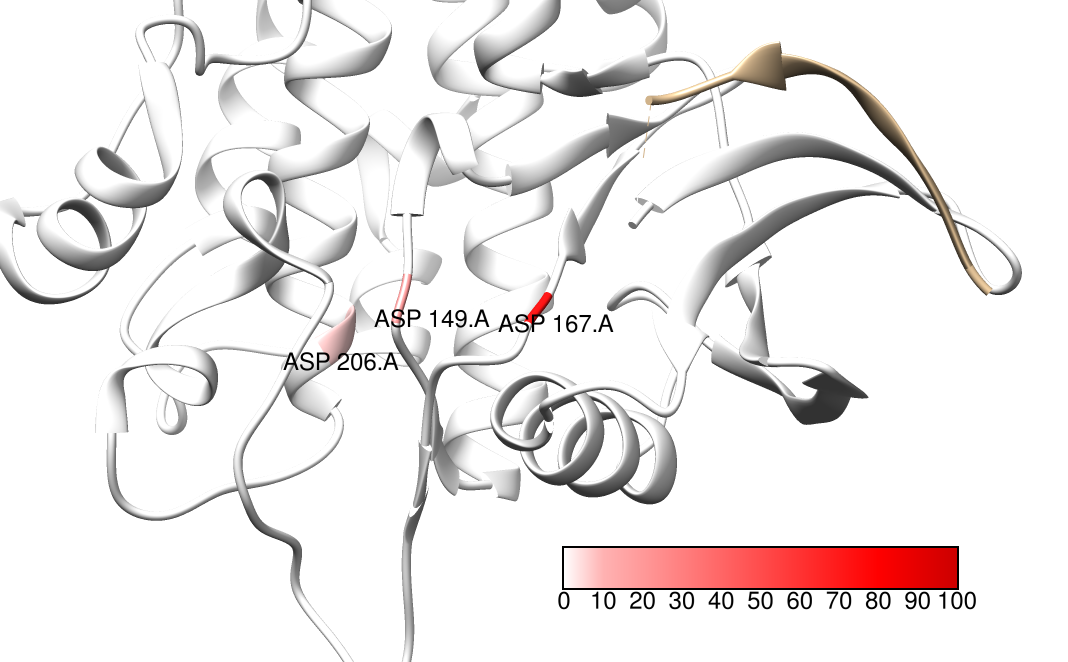}\caption{Attention values at predicted binding sites of MST1 target. The attention
value at residue ASP 149.A is 0.2 (20\%) , 0.72 (72\%) at residue
ASP 167.A, and 0.07 (7\%) at residue ASP 206.A}
\label{fig:mst1_att} 
\end{figure}

\begin{figure}[h]
\centering{}\includegraphics[width=0.8\columnwidth]{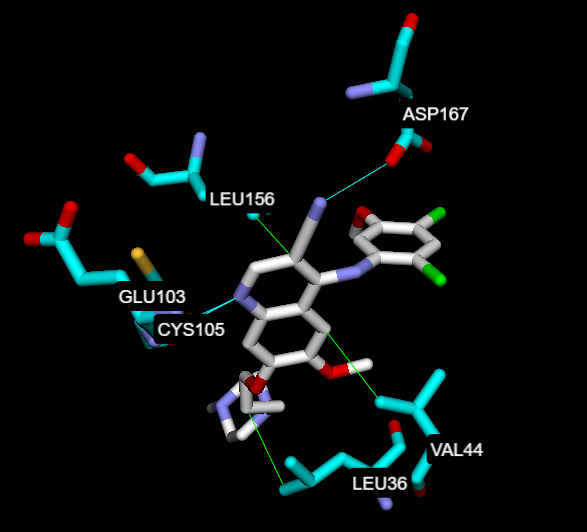}\caption{Residue-ligand interaction between MST1 and Bosutinib predicted by
blind docking.}
\label{fig:mst1_bmode} 
\end{figure}

\begin{figure}[h]
\centering{}\includegraphics[width=0.8\columnwidth]{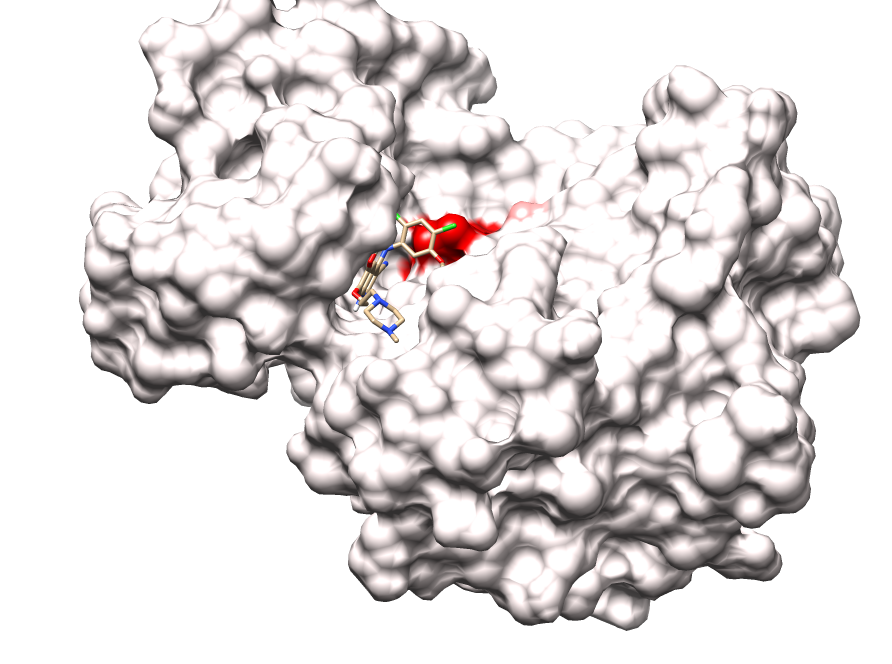}\caption{Predicted binding pocket highlighted in red and the Bosutinib ligand
docking}
\label{fig:surface_mst1} 
\end{figure}

The attention value from the self-attentive layer of the MST1 target
indicates that the residue ASP 167.A has the highest probability (0.72)
to be the binding site. This agrees with the blind docking result.
In the blind docking simulation of the Bosutinib-MST1 pair, the conformation
with the lowest binding energy ($-7.30kcal/mol$) shows the hydrogen
bond between ASP 167.A and the ligand (Fig. \ref{fig:mst1_bmode}).

\begin{figure}[h]
\centering{}\includegraphics[width=0.8\columnwidth]{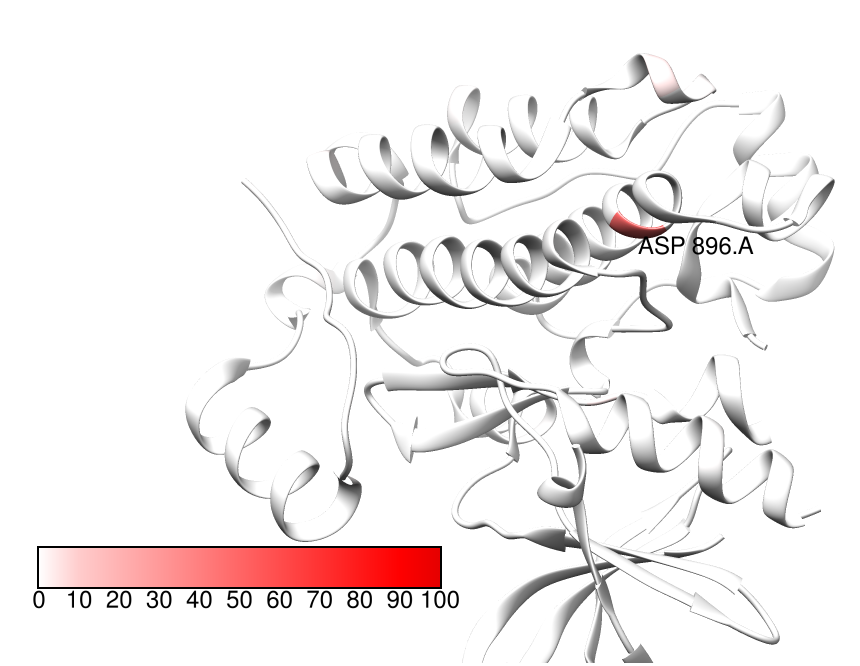}\caption{Attention values at predicted binding sites of EGFR(T790M) target.}
\label{fig:egfr_att} 
\end{figure}

\begin{figure}[h]
\centering{}\includegraphics[width=0.8\columnwidth]{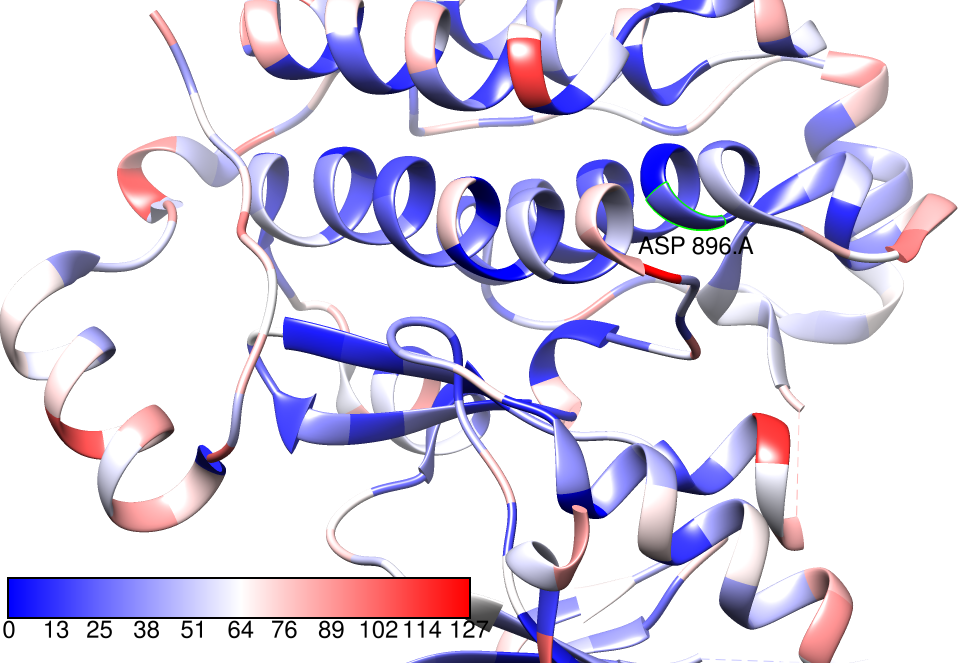}\caption{Solvent-excluded surface areas value of EGFR(T790M) target. }
\label{fig:egfr_ses} 
\end{figure}

\begin{figure}[h]
\centering{}\includegraphics[width=0.8\columnwidth]{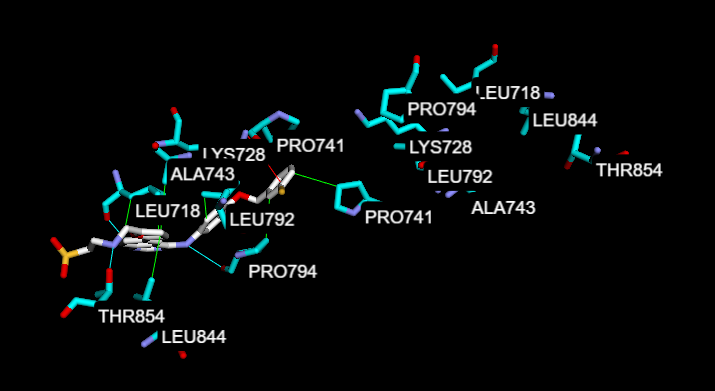}\caption{Residue-ligand interaction between EGFR(T790M) and Lapatinib predicted
by blind docking.}
\label{fig:egfr_dock} 
\end{figure}

For the failed case, we analyze the binding between target EGFR(T790M)
and Lapatinib (PubChem CID 208908). The predicted $K_{D}$ value is
8.7785 while the ground truth value is 6.0655. The attention value
obtained from the self-attentive layer peaks at ASP 896.A with 0.48
(48\%) confidence (see Fig. \ref{fig:egfr_att}). However, in the
crystallised structure of EGFR(T790M), the 896.A residue is buried
(see Fig.\ref{fig:egfr_ses}) and is less likely to be the binding
site. In addition, the predicted conformation of blind docking shows
no interaction between residue 896.A and the ligand (see Fig. \ref{fig:egfr_dock}).
Therefore, in the early fusion model, the edges between the drug node
and binding site residue are not correct. As a result, the error of
the predicted $K_{D}$ value of Lapatinib-EGFR(T790M) is one of the
highest in the Davis test set. From two examples of a successful case
and failed case, obtaining the correct binding site to form the correct
drug node - residues edge is one of the important aspects of the early
fusion approach.

%% file: sections/conclusion.tex
We have proposed a novel deep learning method, called GEFA (Graph
Early Fusion for binding Affinity prediction) for target-drug affinity
prediction, a crucial task for rapid virtual drug screening and drug
repurposing. To improve the power of protein representation, we use
self-supervised to take advantage of a large amount of unlabeled target
sequences. To address the latent representation change due to conformation
change during the binding process, the early fusion between drug and
target is proposed. Unlike the late fusion approach extracting representation
separately, the early fusion approach integrates drug representation
info into protein representation learning phase. The self-attention
value of the target sequence is used as edge weight connecting drug
node and residue node in the target protein graph. Self-attention
allows the model more interpretable as it shows which residues contribute
to the binding process and the level of contribution of each residue.
The quantitative experiments show that the early fusion approach has
advantages over the late fusion approach. Using the embedding feature
as target node feature has advantages over using one-hot encoding.
Residual block design allows stacking multiple GCN layers for better
learning representation capability.

This work opens room for future investigations. Even though the target
representation change is addressed by adding drug node to the target
graph during the representation learning phase, the conformation change,
which is the residue-residue edge connection change, is not addressed.
If we can learn the edge change, we can express the conformation change
caused by the drug-target binding. In addition, in case the target
protein has multiple binding pocket at different regions, the drug
molecule may only bind at one pocket at one time. However, in our
model, the drug node links to all possible binding sites indicated
by self-attention mask. The binding process modeling will be more
accurate if we can combine drug info into the finding drug-residues
edges process.